\documentclass[conference]{IEEEtran}
\IEEEoverridecommandlockouts
\usepackage{cite}
\usepackage{amsmath,amssymb,amsfonts}
\usepackage{algorithmic}
\usepackage{graphicx}
\usepackage{textcomp}
\usepackage{xcolor}

\usepackage{color}
\usepackage{amssymb}
\usepackage{multirow}

\def\BibTeX{{\rm B\kern-.05em{\sc i\kern-.025em b}\kern-.08em
		T\kern-.1667em\lower.7ex\hbox{E}\kern-.125emX}}

\begin{document}
	
\title{PGD: A Large-scale Professional Go Dataset for Data-driven Analytics\\
\thanks{https://github.com/Gifanan/Professional-Go-Dataset}
}

\author{\IEEEauthorblockN{Yifan Gao}
	\IEEEauthorblockA{\textit{School of Biomedical Engineering, Division of Life Science and Medicine} \\
		\textit{University of Science and Technology of China}\\
		Hefei, China \\
		yifangao@stumail.neu.edu.cn}
}

\maketitle

\begin{abstract}
	Lee Sedol is on a winning streak—does this legend rise again after the competition with AlphaGo? Ke Jie is invincible in the world championship—can he still win the title this time? Go is one of the most popular board games in East Asia, with a stable professional sports system that has lasted for decades in China, Japan, and Korea. There are mature data-driven analysis technologies for many sports, such as soccer, basketball, and esports. However, developing such technology for Go remains nontrivial and challenging due to the lack of datasets, meta-information, and in-game statistics. This paper creates the Professional Go Dataset (PGD), containing 98,043 games played by 2,148 professional players from 1950 to 2021. After manual cleaning and labeling, we provide detailed meta-information for each player, game, and tournament. Moreover, the dataset includes analysis results for each move in the match evaluated by advanced AlphaZero-based AI. To establish a benchmark for PGD, we further analyze the data and extract meaningful in-game features based on prior knowledge related to Go that can indicate the game status. With the help of complete meta-information and constructed in-game features, our results prediction system achieves an accuracy of 75.30\%, much higher than several state-of-the-art approaches (64\%-65\%). As far as we know, PGD is the first dataset for data-driven analytics in Go and even in board games. Beyond this promising result, we provide more examples of tasks that benefit from our dataset. The ultimate goal of this paper is to bridge this ancient game and the modern data science community. It will advance research on Go-related analytics to enhance the fan experience, help players improve their ability, and facilitate other promising aspects. The dataset will be made publicly available.
\end{abstract}


\section{Introduction}
Go (baduk, weiqi) is one of the most popular board games in Asia, especially in China, Japan, and Korea \cite{togo}. The attractiveness of Go has grown rapidly in the last two decades. The prize money for professional Go tournaments has reached millions of dollars each year, with dozens of millions of audiences watching competitions from TV streams and online servers. Professional Go players have been using intuition-driven performance analysis methods to improve themselves and prepare for tournaments for a long time. For example, players will replay and review their opponent's matches, realize their shortcomings, and expect to gain an advantage in upcoming games.

\begin{figure}
	\centering
	\includegraphics[width=3.3in]{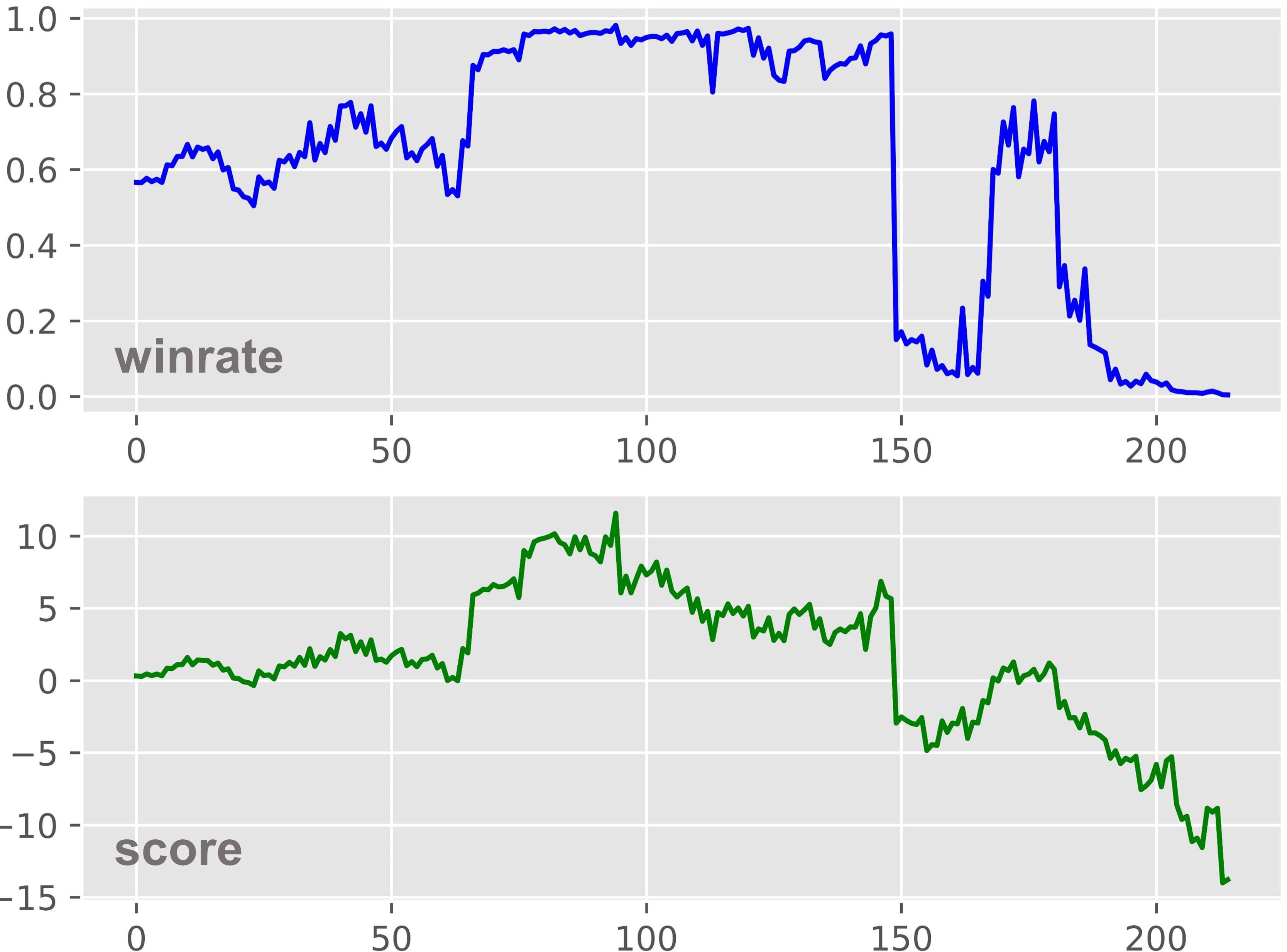}
	\caption{An illustration of KataGo's analysis of a game. The blue line represents the win rate, and the green line represents the score differential. Both are in black's perspective.}
	\label{FIG:katago}
\end{figure}

With the advent of open-source AI based on the AlphaZero \cite{alphazero} algorithm, such as ELF OpenGo \cite{elfopengo}, Leela Zero \cite{leelazero}, and KataGo \cite{katago}, players can improve performance analysis by obtaining in-game statistics from the AI during the game review. Figure \ref{FIG:katago} shows the results of KataGo's analysis of a game. However, these analysis methods are still intuition-driven rather than data-driven, as players rarely use purposefully collected data to conduct further research.

There are many data-driven performance analysis technologies for many sports. Castellar~\cite{tabletennis}  examines the effects of response time, reaction time, and movement time on the performance of table tennis players. Merhej~\cite{happennext} uses deep learning methods to evaluate the value of players' defensive actions in football matches. Baboota~\cite{premierleague} predicts the outcome of English Premier League matches using machine learning methods and achieves promising results. Beal~\cite{combining} uses natural language processing techniques to blend statistical data with contextual articles from human sports journalists and improve the performance of predicting soccer matches. Yang~\cite{moba} proposes an interpretable two-stage Spatio-temporal network for predicting the real-time win rate of mobile MOBA games.

These analysis technologies play an important role in enhancing the fan experience and evaluating player level. However, there is no related technology in Go yet because 1) there is no complete dataset of professional Go games, 2) Go matches do not have the statistic that can indicate the status of both sides (e.g., rebounds in basketball), and 3) compared to popular sports, very few people understand Go, so it isn't easy to organize data and construct effective features well.

In this background, we present the Professional Go Dataset (PGD), containing 98,043 games played by 2,148 professional players. The raw records of the games are derived from publicly available Go datasets, the meta-information related to players and tournaments is annotated by a knowledgeable Go fan, and KataGo analyzes the rich in-game statistics. Thus, the quality and coverage of the dataset are reliable. In addition to this, we further extract and process the in-game statistics through the prior knowledge of Go.

To establish baseline performance on PGD, we used several popular machine learning methods for game outcome prediction. Prediction of Go game results is a challenging performance analysis problem, which has been mainly done by designing rating systems for predicting outcomes. However, the various methods designed have produced rather limited performance improvements. Even the state-of-the-art Bayesian approach, WHR \cite{whr}, has reached a ceiling in terms of accuracy because it utilizes only win-loss and time information. 

The experimental results show a 9.6\% improvement in the accuracy of the prediction system with the help of multiple metadata features and in-game features. In addition to this, we give more promising tasks that can benefit from PGD.

Our contributions are concluded as follows:
\begin{itemize}
	\item We present the first professional Go dataset for sports analytics. The dataset contains a large amount of player, tournament, and in-game data, facilitating extensive performance analysis.
	
	\item We have made the first attempt to feature engineer the statistics within the Go game and developed a machine learning model to predict the outcome of the game. The proposed model significantly outperforms previous rating-based methods.
	
	\item This paper presents possible research directions that benefit from this dataset.
\end{itemize}

\section{Related Work}
\subsection{Background of Professional Go}
Although Go originated in China, the professional Go system first appeared in Japan. During the Edo period, Go became a popular game in Japan and was financed by the government. in the early 20th century, Japan first established the modern professional Go system and maintained its absolute leadership until the 1980s. In 1985, Nie Weiping achieved an incredible streak of victories against top Japanese players in the Japan-China Super Go Tournament. Four years later, Cho Hunhyun defeated Nie Weiping by 3:2 in a highly anticipated final and won the 400,000 dollars championship prize of the Ing Cup. These two landmark events represent the formation of a triple balance of power between China, Japan, and Korea. The golden age of Korean Go began in 1996 and lasted for ten years. Korean players represented by Lee Changho dominated the competition during this period. After that, a large number of Chinese players progressed rapidly and competed fiercely with Korean players, leading to a lot of discussion among fans about the strongest player like Gu Li versus Lee Sedol and Ke Jie versus Shin Jinseo.

Founded in 1988, the Ing Cup and the Fujitsu Cup (discontinued in 2011) were two of the earliest Go world tournaments. Like the Grand Slam in tennis, the world champions of Go are the highest honor for professional players. Table \ref{tab:championship} summarizes the major Go tournaments and their latest champions.

\begin{table}
	\centering
	\renewcommand\arraystretch{1.2}
	\caption{Active major tournaments and their latest champions. Note that the regional tournaments are only partially listed. Text Marked with an asterisk indicates that this is a team tournament (win-and-continue competition). Statistics are as of December 31, 2021.}
	\begin{tabular}{ll}
		\hline
		Tournament Name (Count)      & Latest Champion                                   \\ \hline
		\textbf{World Major}         &                                                   \\
		Samsung Cup (26)             & Park Junghwan 2:1 Shin Jinseo         \\
		LG Cup (26)                  & Shin Minjun 2:1 Ke Jie                \\
		Nongshim Cup* (23)            & Shin Jinseo 1:0 Ke Jie    \\
		Chunlan Cup (13)             & Shin Jinseo 2:0 Tang Weixing          \\
		Ing Cup (9)                  & Tang Weixing 3:2 Park Junghwan        \\
		Mlily Cup (4)                & Mi Yuting 3:2 Xie Ke                  \\
		Bailing Cup (4)              & Ke Jie 2:0 Shin Jinseo                \\
		\hline
		\textbf{Regional Major}      &                                                   \\
		Asian TV Cup (31)            & Shin Jinseo 1:0 Ding Hao               \\
		Japanese Kisei (47)          & Iyama Yuta 4:1 Kono Rin                           \\
		Japanese Meijin (47)         & Iyama Yuta 4:3 Ichiriki Ryo                       \\
		Japanese Honinbo (73)        & Iyama Yuta 4:3 Shibano Toramaru                   \\
		Chinese Mingren (32)         & Mi Yuting 2:1 Xu Jiayang                          \\
		Chinese Tianyuan (35)        & Gu Zihao 2:1 Yang Dingxin                         \\
		Korean Myeongin (44)         & Shin Jinseo 2:1 Byun Sangil                       \\
		Korean KBS Cup (33)          & Shin Jinseo 2:0 An Sungjoon                       \\
		Taiwan Tianyuan (20)         & Wang Yuanjun 4:2 Jian Jingting                    \\
		European Go Grand Slam (5)   & Mateusz Surma 1:0 Artem Kachanovsky               \\ \hline
	\end{tabular}
	\label{tab:championship}
\end{table}

\subsection{State-of-the-art Go AI in the Post-AlphaZero Era}
AlphaZero has outperformed the best professional players by a wide margin and has prompted researchers to turn their attention to more challenging tasks in AI. However, in the Go community, AlphaZero and its open-source implementations, like ELF OpenGo, early Leela Zero, do not work well for analyzing human games: On the one hand, their evaluation contains only the win rate and not the score difference. In game states where the gap is small, the AI's win rate fluctuates dramatically, so the robustness is poor. On the other hand, these AIs cannot support different rules and \textit{komis} (additional points added for the white player to compensate for the black player's first move advantage), thus often leading to incorrect evaluation of games.

KataGo is the cutting-edge AlphaZero-based algorithm that can provide rich in-game statistics while analyzing more accurately. In addition to that, KataGo supports different rules and komis. Therefore, KataGo is suitable for analyzing human games and getting in-game statistics. While KataGo supports a variety of statistics, in the following section, we mainly focus on KataGo's three main stats: win rate, score difference, and recommended move list. 

\subsection{Board Game Analytics}
Many board games have AI based on the AlphaZero algorithm, such as Gomoku  \cite{gomokunet}, Othello \cite{othello}, and NoGo \cite{nogo}. However, these games do not have a large number of accessible game records, which makes it challenging to develop analysis techniques for them.

On the other hand, the study of chess analytics emerged very early, with the benefit of detailed chess datasets, a large number of participants, and advanced chess computer engines. The most common applications contain the evaluation of the player's rating \cite{intrinsic,actualplay,evolutionaryrating}, skill \cite{skillperformance,skillcomputer}, and style \cite{preference,psychology,psychometric}.

These efforts have made a significant contribution to the chess community. However, due to the lack of benchmarks and discontinuous research, the AI community and the game analytics community have rarely focused on chess-related research, especially in light of the recent rapid development of machine learning methods. Acher~\cite{largescale} has attempted to establish a large-scale open benchmark for chess, but the project seems to be inactive at the moment. Fadi~\cite{resultprediction} introduces machine learning methods into game result prediction. Still, the effect appears to be unsatisfactory, as the performance of the approaches based on machine learning models is lower than the Elo rating system. Recent research has made significant progress in modeling human behavior in chess \cite{mcilroy2020aligning, mcilroy2020learning, mcilroy2021detecting}. In particular, \cite{mcilroy2021detecting} has developed a technology to determine a player's identity from a sequence of chess moves. It demonstrates the great potential of data-driven board game analytics.

There is little previous research related to Go analytics. However, the rapid growth in the number of Go tournaments, players, and recorded games in the last two decades has made it possible to develop a dataset and benchmark for data-driven Go analytics. Therefore, this paper proposes PGD, a large-scale professional Go dataset, for data-driven sports analytics. We hope to build a bridge between this ancient game and modern game data science with this dataset.

\begin{table}
	\centering
	\renewcommand\arraystretch{1.2}
	\caption{Regions where both players are located. Legend: CR = Cross Region}
	\begin{tabular}{cccccc}
		\hline
		Total & CR    & CHN   & KOR   & JPN   & Others \\
		98043 & 12963 & 26977 & 19455 & 32907 & 5741    \\ \hline
	\end{tabular}
	\label{tab:regions}
\end{table}

\begin{figure}
	\centering
	\includegraphics[width=3.3in]{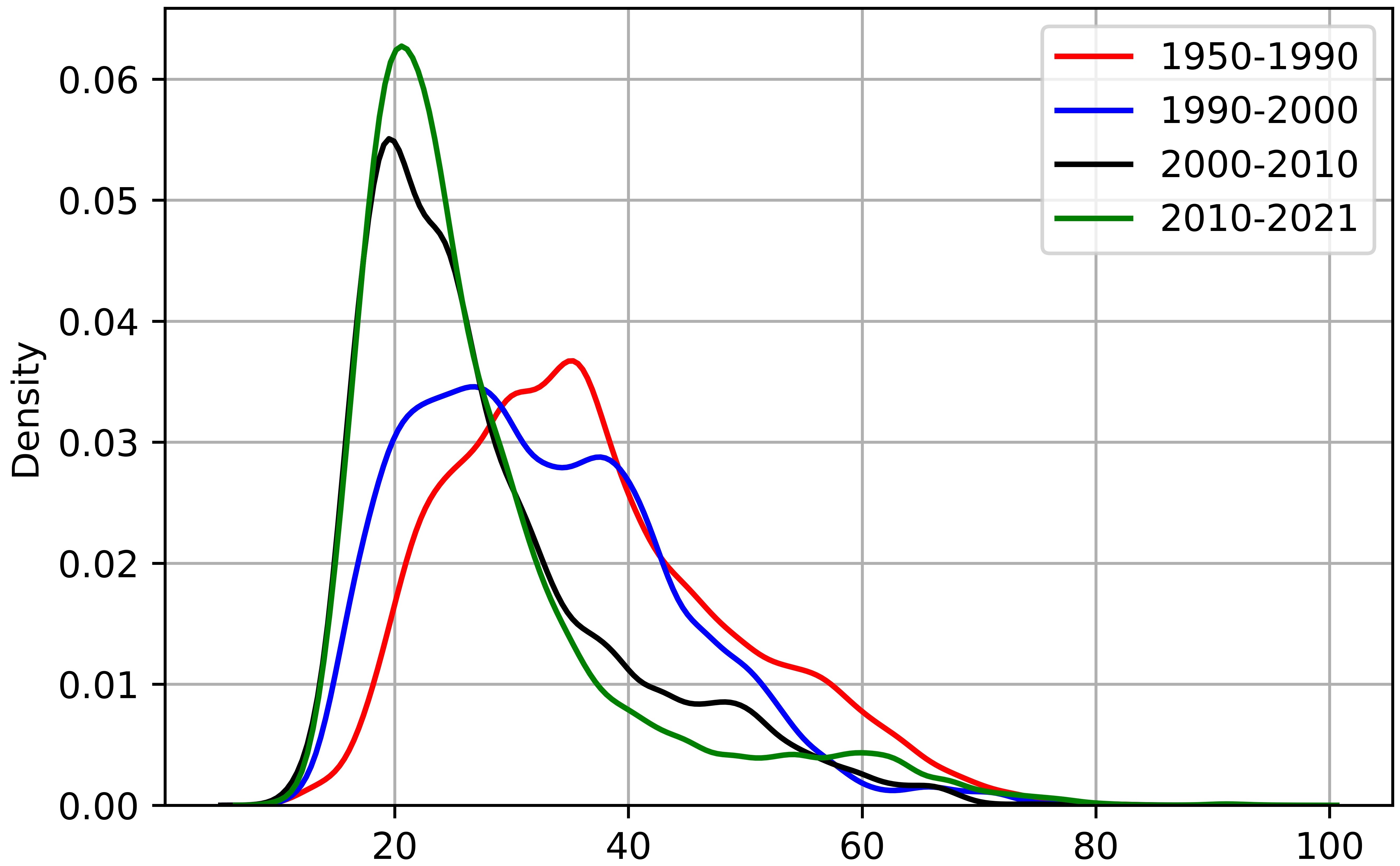}
	\caption{Distribution of the age of players in different generations.}
	\label{FIG:age}
\end{figure}

\section{Dataset Collection and Annotation}
\subsection{Meta-information}
\label{sec:meta}
First, we access the raw game data through Go4Go \footnote{https://www.go4go.net}, the largest database of Go game records. The owner approved the academic use of the database. In the raw data, we excluded some games, including amateur or AI competitions, handicap games, and games that ended abnormally (e.g., lose by forfeit after playing a \textit{ko} without the \textit{ko threat}). In the end, we obtained 98,043 games played by 2,148 professional players from 1950 to 2021 in SGF format.

We obtain information about each player's date of birth, gender, and nationality in Sensei's Library\footnote{https://senseis.xmp.net} and List of Go Players\footnote{https://db.u-go.net}. If the player's metadata is not in the database, a high-level amateur go player would supplement the information. We set the metadata to "unknown" if the player could not make an affirmative judgment.

Meta-information related to matches and rounds is extracted from SGF files. We finally got 503 tournament categories (e.g., Ing Cup, Samsung Cup), 3131 tournaments (e.g., 1st Ing Cup, 15th Samsung Cup), and 342 rounds (e.g., round 1, semi-finals). The amateur player manually labels each tournament, including the tournament's region, importance, and tournament type. The region includes international and non-international; the importance includes world-major and regional-major; the types of tournaments include elimination, league, team, and friendly.

\subsection{In-game Statistics}
We used KataGo v1.9.1 with tensorRT backend to analyze the game records. We applied the following strategy to ensure a balance of accuracy and speed: 100 simulations were performed for each move to obtain initial in-game statistics. If a move resulted in a large fluctuation (more than 10\% win rate or 5 points), KataGo would re-evaluate the action with a simulation count of 1000. In the end, we obtained in-game statistics for all games, mainly including win rate, score difference, and preferred moves. Our analysis was conducted on an NVIDIA RTX 2080Ti graphic card and took a total of about 25 days.

\section{Statistics}
In this section, we present the statistics of PGD. The statistics include two aspects: players and games.
\subsection{Players}

\begin{table}
	\centering
	\renewcommand\arraystretch{1.2}
	\caption{Most frequent players.}
	\begin{tabular}{cc|cc}
		\hline
		Players           & Games & Players           & Games  \\ \hline
		Cho Chikun       & 2079       & O Rissei         & 1217        \\
		Lee Changho      & 1962       & Yamashita Keigo  & 1217        \\
		Kobayashi Koichi & 1605       & Gu Li            & 1215       \\
		Rin Kaiho        & 1536       & Otake Hideo      & 1214       \\
		Cho Hunhyun      & 1533       & Cho U            & 1145       \\
		Lee Sedol        & 1375       & Choi Cheolhan    & 1133       \\
		Yoda Norimoto    & 1316       & Chang Hao        & 1100       \\
		Kato Masao       & 1249       & Park Junghwan    & 1058       \\
		Takemiya Masaki  & 1248       & Kobayashi Satoru & 1044       \\ \hline
	\end{tabular}
	\label{tab:frequentplayers}
\end{table}

\begin{table}
	\centering
	\renewcommand\arraystretch{1.2}
	\caption{Most frequent matchups.}
	\begin{tabular}{cccc}
		\hline
		Matchups                        & Games & Win & Loss \\ \hline
		Cho Hunhyun vs. Lee Changho     & 287   & 110 & 177  \\
		Cho Hunhyun vs. Seo Bongsoo     & 207   & 139 & 68   \\
		Kobayashi Koichi vs. Cho Chikun & 123   & 60  & 63   \\
		Yoo Changhyuk vs. Lee Changho   & 122   & 39  & 83   \\
		Cho Chikun vs. Kato Masao       & 107   & 67  & 40   \\ \hline
	\end{tabular}
	\label{tab:frequentmatchups}
\end{table}

\begin{table}
	\centering
	\renewcommand\arraystretch{1.2}
	\caption{Most frequent tournaments.}
	\begin{tabular}{cc|cc}
		\hline
		Tournaments       & Games & Tournament             & Games \\ \hline
		Chinese League A & 11092 & Japanese Oza           & 1996  \\
		Korean League A  & 4773  & Japanese NHK Cup       & 1939  \\
		Japanese Honinbo & 3655  & Samsung Cup            & 1806  \\
		Japanese Ryusei  & 2932  & Chinese Mingren        & 1588  \\
		Japanese Meijin  & 2869  & Chinese Women's League & 1569  \\
		Japanese Judan   & 2766  & LG Cup                 & 1514  \\
		Japanese Kisei   & 2723  & Korean League B        & 1411  \\
		Japanese Tengen  & 2383  & Chinese Tianyuan       & 1282  \\
		Japanese Gosei   & 2230  & Korean Women's League  & 1279  \\ \hline
	\end{tabular}
	\label{tab:tournaments}
\end{table}

\begin{table}
	\centering
	\renewcommand\arraystretch{1.2}
	\caption{Black win rate (BWR) under different komis.}
	\begin{tabular}{cccc}
		\hline
		Komis & Games & KataGo BWR & Actual BWR \\ \hline
		4.5   & 1123  & 64\%       & 55.48\%    \\
		5.5   & 21980 & 56\%       & 53.20\%    \\
		6.5   & 46952 & 48\%       & 50.43\%    \\
		7.5   & 26819 & 39\%       & 46.90\%    \\
		8     & 1169  & 39\%       & 48.85\%    \\ \hline
	\end{tabular}
	\label{tab:komis}
\end{table}

\begin{figure*}
	\centering
	\includegraphics[width=7in]{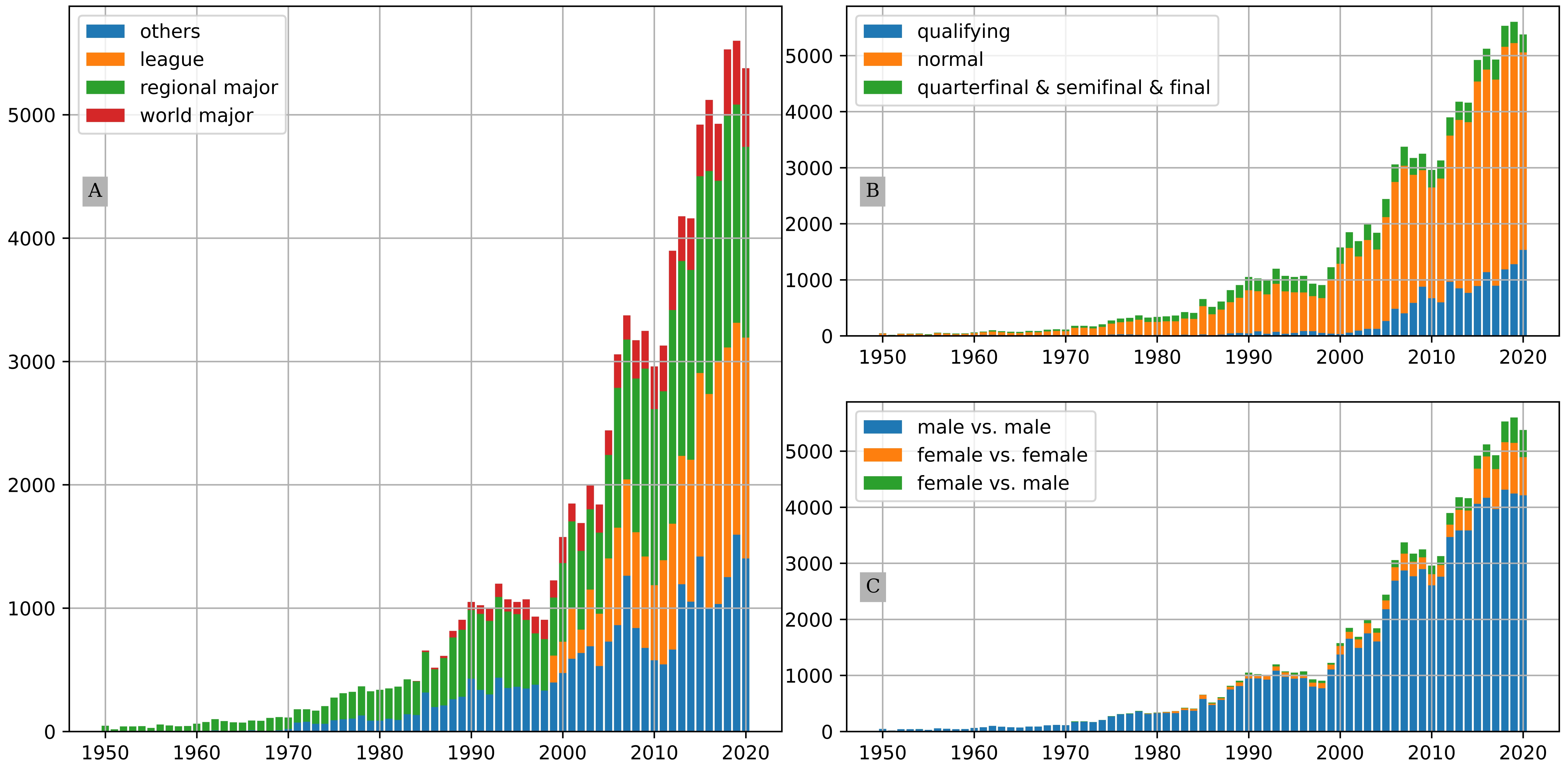}
	\caption{Game counts in years.}
	\label{FIG:games}
\end{figure*}

\begin{figure}
	\centering
	\includegraphics[width=3.3in]{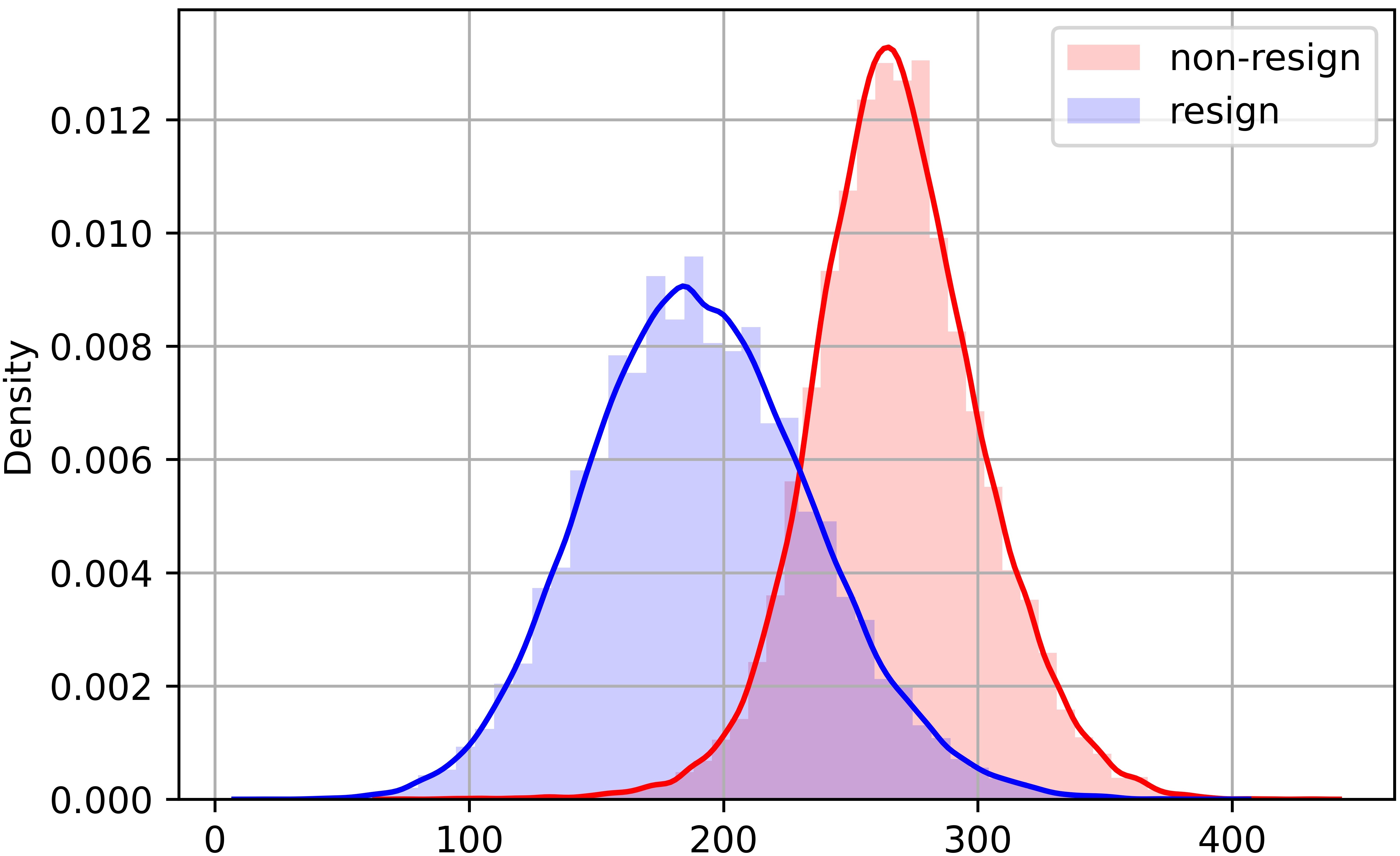}
	\caption{Distribution of game lengths.}
	\label{FIG:length}
\end{figure}

Table \ref{tab:regions} summarizes the diversity of the competitions, and we provide the regions where both sides of the games are located. It can be seen that most games where both players are from Japan, while there are fewer cross-region games. Figure \ref{FIG:age} presents the age distribution of the players in the different generations of the games. Before 1990, the age of the players was between 30-50 years old. In the following decade, the average age of players dropped rapidly to less than 30 years old. In the 21st century, players in their 20s completed the most competitions. This trend reveals that professional Go is becoming more competitive. Table \ref{tab:frequentplayers} and Table \ref{tab:frequentmatchups} show the most frequent players and matchups. These prominent legends completed most of the games, indicating a long-tailed distribution of the number of games played by different players.

\subsection{Games}

\begin{figure}
	\centering
	\includegraphics[width=3.3in]{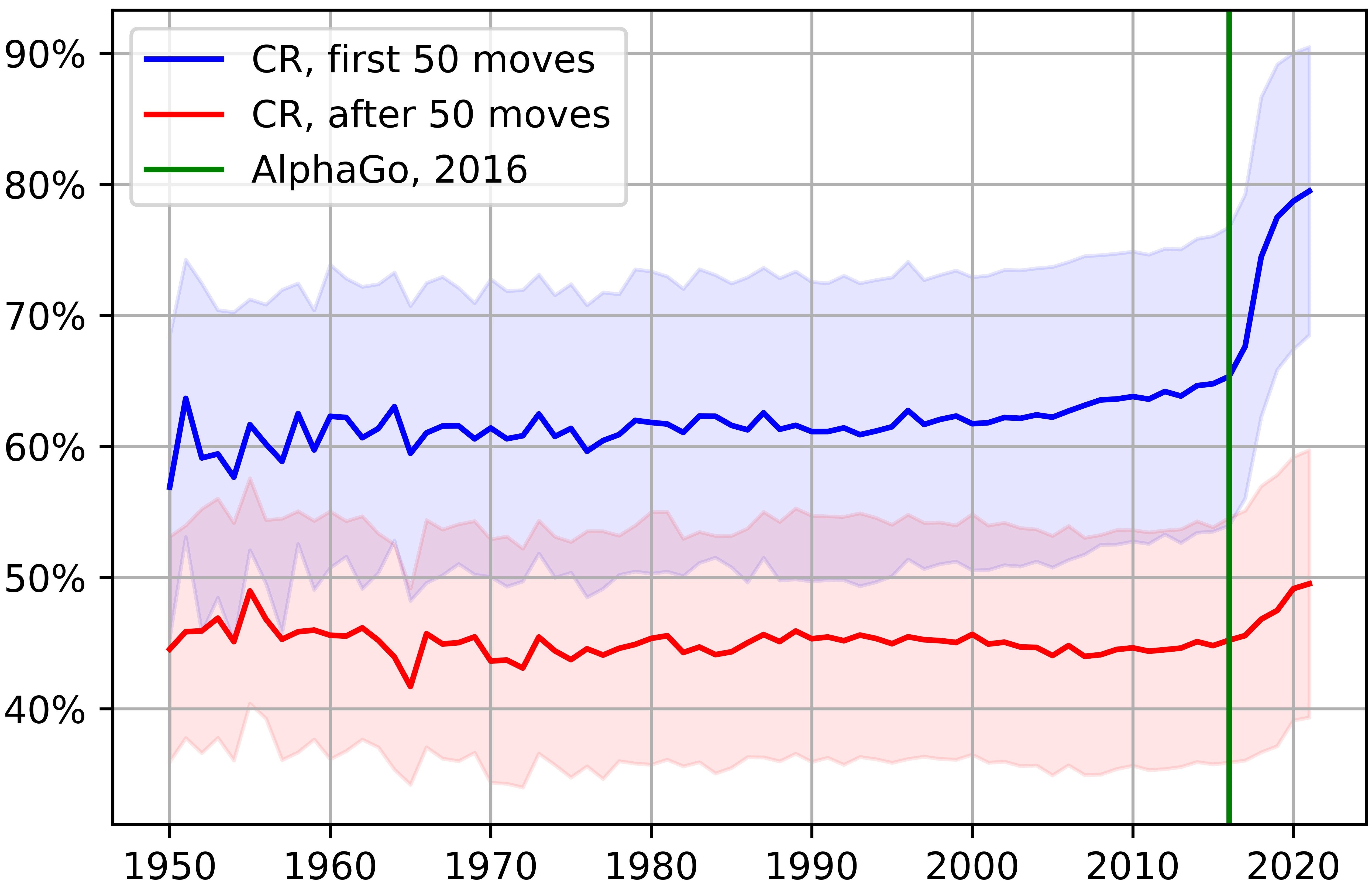}
	\caption{Mean coincidence rate (CR) in years.}
	\label{FIG:coin}
\end{figure}

\begin{figure}
	\centering
	\includegraphics[width=3.0in]{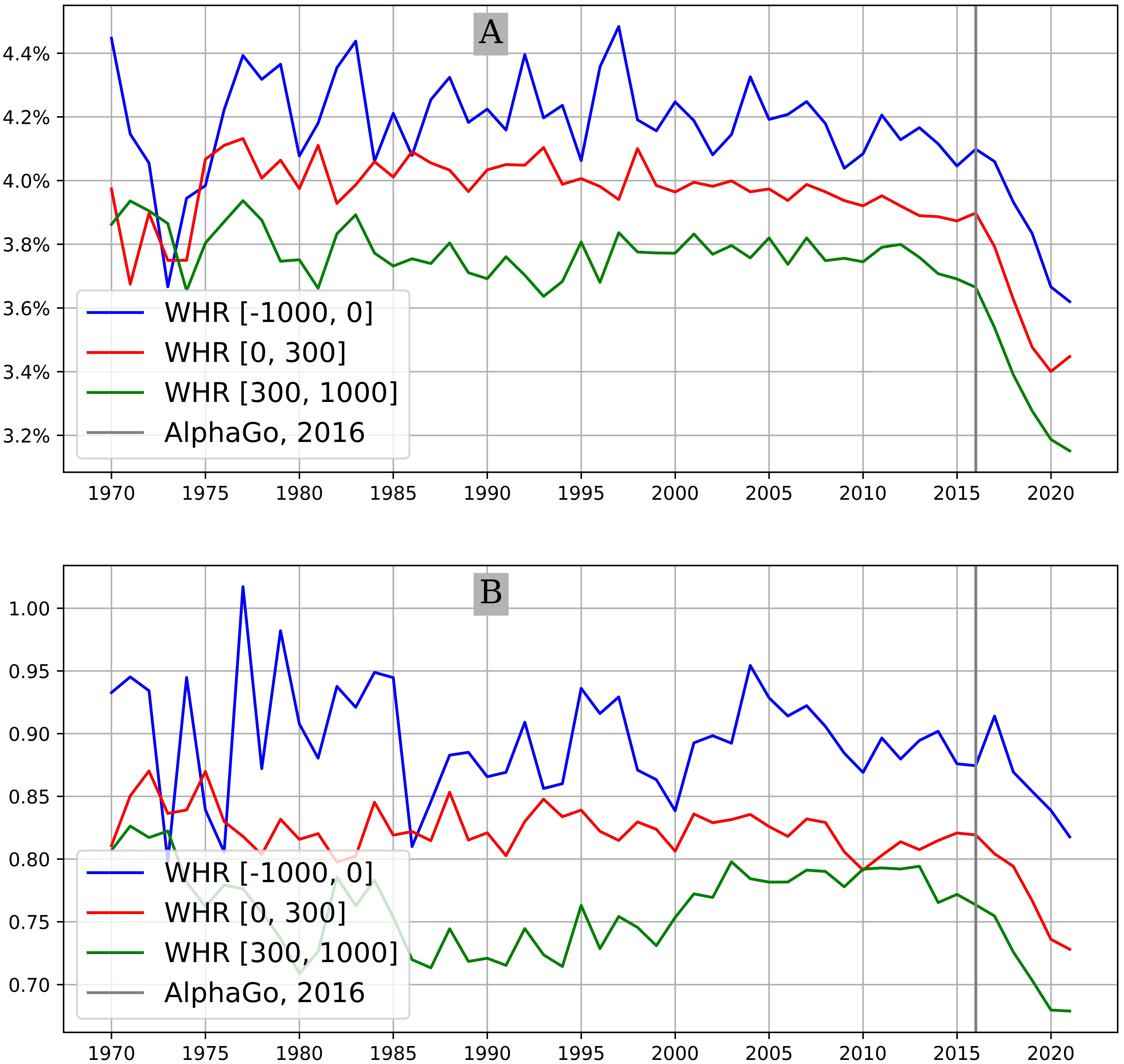}
	\caption{Mean loss win rate (MLWR) and mean loss score (MLS) with different WHR score in years.}
	\label{FIG:loss}
\end{figure}

Figure \ref{FIG:games} presents the number of games per year in the PGD. We can observe it grows exponentially. On the one hand, it shows the increase in the number of professional matches; on the other hand, it shows that PGD did not record many early games. Figure \ref{FIG:games}A shows the number of different types of tournaments. We can see that the creation and development of leagues in the 21st century have provided many opportunities for Go players. Figure \ref{FIG:games}B shows the round statistics of the matches. Figure \ref{FIG:games}C shows the gender statistics of the games. We can observe that the number of tournaments for female professional Go players is increasing but still needs further improvement.

Table \ref{tab:tournaments} shows the tournaments with the highest number of games, which are mostly Japanese tournaments, except for the Chinese League A and the Korean League A. This is mainly because these tournaments were held for many years. Table \ref{tab:komis} shows the win rate of black in different komis. The win rate for black players is 9\% higher in the 4.5 komis than 7.5 komis. Thus, the different komis shift the distribution of statistics within many games, creating additional difficulties in developing prediction techniques. Figure \ref{FIG:length} shows the distribution of game lengths, where the lengths of resigned games are significantly shorter.

Figure \ref{FIG:coin} and Figure \ref{FIG:loss} show the trends of some in-game statistics under different years. From these schematics, we observe some interesting results. First, the advent of AlphaGo prompted professional players to start imitating the AI's preferred moves, which is particularly evident in the opening phase (first 50 moves), and one can observe the coincidence rate in Figure \ref{FIG:coin}, which grows very significantly after 2016. At the same time, the non-opening phase is difficult to imitate, so the growth of the coincidence rate is much smaller than in the opening stage. From Figure \ref{FIG:loss}, We can observe that the average loss of win rate and the average loss of score also decreased rapidly after the emergence of AlphaGo. It is worth noting that Figure \ref{FIG:loss} also shows statistics for players with different WHR ratings, with players with higher scores having lower statistics. This offers the potential to develop sports performance analysis techniques.

\section{Game Results Prediction of PGD}
This section develops a system for predicting future matches from historical data. We first perform feature extraction and preprocessing, then apply popular machine learning methods, and finally, we report the performance of the prediction system.

\subsection{Feature Extraction}
In this section, we divide all features into three categories: meta-information features, contextual features, and in-game features. The detailed meanings of these features are explained in detail in the following sections.
\subsubsection{Meta-information Features}
\begin{itemize}
	\item Basic Information (BI): Include the games time, age, gender, region, and other fundamental characteristics of the players.
	\item Ranks (R): Ranking of Go players, 1-dan lowest, 9-dan highest.
	\item WHR Score (WS): WHR rating, which measures the level of the player.
	\item WHR Uncertain (WU): A measure of uncertainty in the WHR rating system. In general, players who have not played for longer or have fewer total games have higher uncertainty.
	\item Tournament Feature (TF): Features of the tournament, already described in Section \ref{sec:meta}.
\end{itemize}

\subsubsection{Contextual Features}
\begin{itemize}
	\item Match Results (MR): The player's recent performance in various competitions, such as 14 wins and 6 losses in the last 20 games, 9 wins and 1 loss in the last 10 games.
	\item Match Results by Region (MRR): Performance against the opponent's region, such as 10 wins and 10 losses in the last 20 games against Korean players.
	\item Matchup Results (MUR): Past competitions against the opponent.
	\item Tournament Results (TR): Past performances at this tournament, such as Ke Jie's 20 wins and 6 losses at the Samsung Cup.
	\item Opponents Ranks (OR): Rank of opponents in recent matches.
	\item Opponents Ages (OA): Age of opponents in recent matches.
	\item Cross-region Counts (CRC): The number of cross-regional competitions. Generally speaking, the larger the number, the higher the level of players.
\end{itemize}

\subsubsection{In-game Features}
\begin{itemize}
	\item Garbage Moves (GM): We define the GM as a move in a game in which the winner is almost determined. GM is calculated as follows: if after move $x$ in a match, under moving average with a window of length 4, the win rate of the leading player is always more than 90\% or the score difference is always more than 3 points, then all moves after step $x$ are called GM. We take the ratio of GM to game lengths as an in-game feature. In the following in-game feature calculations, we remove all GM.
	\item Unstable Rounds (UR): We define the UR as a round in which the win rate or score difference fluctuates dramatically. The UR in the game is calculated as follows: move $x$ and move $x+1$ lose the win rate $w_1$, $w_2$ or score $s_1$, $s_2$ respectively ($w_1$ and $w_2$ are greater than 10\%, $s_1$ and $s_2$ are greater than 5). Also, the absolute value of the difference between $w_1$ and $w_2$ is less than 2\%, or the absolute value of the difference between $s_1$ and $s_2$ is less than 1. The rounds at move $x$ and move $x+1$ are said to be the UR. We take the number of UR as an in-game feature. In the following in-game feature calculations, we remove all UR.
	\item Mean Win Rate (MWR): Average win rate in recent games. Note that the win rate here is the average in-game win rate. For example, Ke Jie comebacks and win in a  game, but the MWR in this game is only 15\%.
	\item Mean Score (MS): Average score difference in recent games.
	\item Mean Loss Win Rate (MLWR): Average win rate lost in recent games.
	\item Mean Loss Score (MLS): Average score lost in recent games.
	\item Advantage Rounds (AR): Number of rounds with a 5\% win rate or a 3 point advantage in recent games.
	\item Strong Advantage Rounds (SAR): Number of rounds with a 10\% win rate or a 5 point advantage in recent games.
	\item Coincidence Rate (CR): The ratio of moves that match KataGo's recommendation to total moves in recent games.
\end{itemize}

\subsection{Model Training}
We selected four advanced machine learning methods to train the models: Random Forest (RF) \cite{randomforest}, XGBoost \cite{xgboost}, LightGBM \cite{lightgbm}, and CatBoost \cite{catboost}. As a comparison, three models based on rating systems were applied to predict game outcomes, including ELO \cite{elo}, TrueSkill \cite{trueskill}, and WHR. It is worth noting that we directly used the default hyperparameters in the corresponding Python package without tuning them. Although adjusting these hyperparameters could further improve the performance, our proposed approach has already achieved very competitive results compared with the other rating system-based models.

We divided the dataset into a training set and a test set. The training set includes competitions from 1950 to 2017, with a total of 77,182 matches. The test set includes competitions from 2018 to 2021, with a total of 20,861 matches. The evaluation metrics use the Accuracy (ACC) and Mean Square Error (MSE).

\begin{table*}[]
	\centering
	\renewcommand\arraystretch{1.2}
	\caption{Experimental results. Bolded font indicates the best performance. The red font in the last row shows the performance difference between our proposed approach and the best rating system-based method.}
	\begin{tabular}{lllllllllllll}
		\hline
		\multicolumn{1}{c}{} & \multicolumn{2}{c}{Mean}                            & \multicolumn{2}{c}{CR}                              & \multicolumn{2}{c}{CHN}                             & \multicolumn{2}{c}{KOR}                             & \multicolumn{2}{c}{JPN}                             & \multicolumn{2}{c}{Others}                          \\ \hline
		\multicolumn{1}{c}{} & \multicolumn{1}{c}{ACC↑} & \multicolumn{1}{c}{MSE↓} & \multicolumn{1}{c}{ACC↑} & \multicolumn{1}{c}{MSE↓} & \multicolumn{1}{c}{ACC↑} & \multicolumn{1}{c}{MSE↓} & \multicolumn{1}{c}{ACC↑} & \multicolumn{1}{c}{MSE↓} & \multicolumn{1}{c}{ACC↑} & \multicolumn{1}{c}{MSE↓} & \multicolumn{1}{c}{ACC↑} & \multicolumn{1}{c}{MSE↓} \\ \hline
		ELO                  & 0.6515                   & 0.2144                   & 0.6466                   & 0.2174                   & 0.6176                   & 0.2273                   & 0.6339                   & 0.2191                   & 0.6637                   & 0.2107                   & 0.7156                   & 0.1907                   \\
		TrueSkill            & 0.6439                   & 0.2605                   & 0.6380                   & 0.2781                   & 0.6095                   & 0.1875                   & 0.6265                   & 0.2654                   & 0.6552                   & 0.2546                   & 0.7106                   & 0.2062                   \\
		WHR                  & 0.6567                   & 0.2125                   & 0.6684                   & 0.2090                   & 0.6212                   & 0.2295                   & 0.6397                   & 0.2193                   & 0.6577                   & 0.2108                   & 0.7254                   & 0.1813                   \\ \hline
		RF                   & 0.6932                   & 0.2022                   & 0.6954                   & 0.1998                   & 0.6623                   & 0.2146                   & 0.6800                   & 0.2086                   & 0.7031                   & 0.1956                   & 0.7446                   & 0.1847                   \\
		XGBoost              & 0.7351                   & 0.1700                   & 0.7457                   & 0.1663                   & 0.7033                   & 0.1844                   & 0.7197                   & 0.1779                   & 0.7563                   & 0.1607                   & 0.7692                   & 0.1520                   \\
		LightGBM             & 0.7509                   & 0.1637                   & 0.7611                   & 0.1574                   & 0.7241                   & 0.1765                   & 0.7374                   & 0.1715                   & 0.7599                   & 0.1600                   & 0.7912                   & 0.1432                   \\
		CatBoost             & \textbf{0.7530}          & \textbf{0.1623}          & \textbf{0.7632}          & \textbf{0.1572}          & \textbf{0.7258}          & \textbf{0.1752}          & \textbf{0.7379}          & \textbf{0.1699}          & \textbf{0.7633}          & \textbf{0.1577}          & \textbf{0.7946}          & \textbf{0.1411}          \\
		& \textcolor{red}{9.6\%}           & \textcolor{red}{-0.050}          & \textcolor{red}{9.5\%}           & \textcolor{red}{-0.052}          & \textcolor{red}{10.5\%}          & \textcolor{red}{-0.052}          & \textcolor{red}{9.8\%}           & \textcolor{red}{-0.049}          & \textcolor{red}{10.0\%}          & \textcolor{red}{-0.053}          & \textcolor{red}{6.9\%}           & \textcolor{red}{-0.040}          \\ \hline
	\end{tabular}
	\label{tab:results}
\end{table*}

\begin{table}
	\centering
	\renewcommand\arraystretch{1.2}
	\caption{Ablation results.}	
	\begin{tabular}{ccccc}
		\hline
		Metadata & Contextual & In-game & ACC↑   & MSE↓   \\ \hline
		\checkmark        &           &        & 0.6719 & 0.1975 \\
		& \checkmark          &        & 0.7099 & 0.1827 \\
		&           & \checkmark       & 0.6883 & 0.1891 \\
		\checkmark        & \checkmark          &        & 0.7342 & 0.1706 \\
		\checkmark        & \checkmark          & \checkmark       & 0.7530  & 0.1623 \\ \hline
	\end{tabular}
	\label{tab:ablation}
\end{table}

\subsection{Experiment Results}
Table \ref{tab:results} shows the experimental results. We can see that WHR achieves an accuracy of 65.67\% and an MSE of 0.2125 among all rating system-based outcome prediction models, which is the best performer. WHR performs slightly better than ELO because WHR takes full advantage of the long-term dependence on game results. The performance of TrueSkill is lower, probably because it is more suitable for calculating the rating of multiplayer sports.

Among the machine learning methods using various features, CatBoost has the highest performance, reaching an accuracy of 75.30\% and an MSE of 0.1623, which is much higher than WHR. In almost every category, the CatBoost model brings a 10\% improvement in accuracy and an MSE improvement of 0.05. The improvement of the Others category is lower because it is relatively easier to predict, and the accuracy of the WHR method exceeds 72\%.

\subsection{Ablation Study}
We also designed ablation experiments to verify the validity of each modular feature, each using the CatBoost method. The results are shown in Table \ref{tab:ablation}. The model using only meta-information features is 67.19\%, which is slightly higher than the WHR method, indicating that other attributes in meta-information besides WHR also contribute to the results. The accuracy of 70.99\% was achieved using only contextual features. In comparison, the accuracy of 68.83\% was achieved by using only in-game features, both higher than the state-of-the-art rating system-based prediction approach. The combination of these features further improves the predictive ability of the model. The ablation experiments demonstrate that the characteristics of each of our modules enhance the performance of the prediction system.

\section{Possible Research Directions}
This section will focus on some of the research directions that benefit from PGD, in addition to game outcome prediction.
\subsection{Feature engineering of in-game statistics} Although our extracted in-game features can analyze player performance better, there is still room for improvement. First, more features could be designed to enhance the predictive ability. Second, as seen in Figure \ref{FIG:katago}, well-developed time series techniques have the potential to improve the analysis.

\subsection{Behavior and style modeling} 
The behavior and style of different players attract the most attention of fans in Go matches. For example, Lee Changho's style is extreme steadiness and control, overcoming his opponent at the last moment. On the other hand, Gu Li is characterized by taking every opportunity to fight with his opponent.

Modeling and classifying the playing styles of players is a challenging but fascinating problem. It can give many potential applications, such as targeted training and preparations, AI-assisted cheating detection \cite{cheating1,cheating2}, even wide-range psychological research \cite{psychology,psychometric,risk,female}. 

There is a relatively small amount of literature on this problem in board games. Omori~\cite{shogi} proposes classifying shogi moves based on game style and training AI for a specific game style. McIlroy-Young \cite{mcilroy2020learning} develops a personalized model that predicts a specific player's move and demonstrates that it captures individual-level human playing styles. Style modeling and recognition have not made impressive success due to the lack of proper analysis techniques and large-scale datasets. The advent of PGD has helped to make progress on such issues.

\subsection{Rating system}
The result prediction model can only give the winning percentage between two players. On the other hand, a rating system can incorporate the relative strengths of many players into the evaluation and is more valuable in assessing a player's strength. Existing rating systems for board games only consider win-loss and time information, while a rating system that incorporates other features has the potential to evaluate a player's strength better. Thus the combination of machine learning approaches and traditional rating systems is promising.

\subsection{Live commentary enhancement}
Millions of Go fans watch Go matches live on online platforms or TV channels every year. However, a classic world Go tournament can last more than five hours with only the host analyzing the game. In this case, viewers get bored and stop watching. 

With the advent of AlphaZero, live Go broadcasts usually show the AI's evaluation of the position, which slightly improves the viewer experience. However, simply showing the win rate is not enough to attract viewers to watch for a long time and often leads to attacks on professional players, as even the top professional players' games are usually rated very negatively by the AI.

Our proposed PGD has the potential to change this situation. A range of technologies developed by PGD, like automatic commentary text generation, real-time statistics, and result prediction, will enable a much more detailed and nuanced analysis of Go matches, thus greatly enhancing the viewer experience in live streaming.

\section{Conclusion}
We present PGD, the first large-scale professional Go dataset for data-driven analytics. With a large number of valuable features, our dataset can be used as a benchmark for a wide range of performance analysis tasks related to professional Go. The results show that machine learning methods far outperform state-of-the-art rating systems-based approaches in game results prediction. Our dataset is promising for developing useful tools and solving real-world problems for professional Go systems.

\bibliographystyle{IEEETran}
\bibliography{conference_101719}
	
\end{document}